# Utilizing Machine Learning Models to Predict Acute Kidney Injury in Septic Patients from MIMIC-III Database

Aleyeh Roknaldin [a], Zehao Zhang [a], Jiayuan Xu [a], Kamiar Alaei [b], and Maryam Pishgar [a]

[a] Department of Industrial and Systems Engineering
University of Southern California, Los Angeles, CA 90007, United States of America
[b] Department of Health Science
California State University, Long Beach, 1250 Bellflower Blvd. HHS2-117, Long Beach, CA 90840, USA

Email Addresses of Authors:
Roknaldin: roknaldi@usc.edu, Zhang: zhangzeh@usc.edu, Xu: jiayuanx@usc.edu, Alaei: kamiar.alaei@csulb.edu , Pishgar: pishgar@usc.edu

## Abstract

Background: Sepsis is a severe condition that causes the body to respond incorrectly to an infection. This reaction can subsequently cause organ failure, a major one being acute kidney injury (AKI). For septic patients, approximately 50% develop AKI, with a mortality rate above 40%. Creating models that can accurately predict AKI based on specific qualities of septic patients is crucial for early detection and intervention.

Methods: Using medical data from septic patients during intensive care unit (ICU) admission from the Medical Information Mart for Intensive Care 3 (MIMIC-III) database, we extracted 3301 patients with sepsis, with 73% of patients developing AKI. The data was randomly divided into a training set (n = 1980, 40%), a test set (n = 661, 10%), and a validation set (n = 660, 50%). The proposed model was logistic regression, and it was compared against five baseline models: XGBoost, K Nearest Neighbors (KNN), Support Vector Machines (SVM), Random Forest (RF) and LightGBM. Area Under the Curve (AUC), Accuracy, F1-Score, and Recall were calculated for each model.

Results: After analysis, we were able to select 23 features to include in our model, the top features being urine output, maximum bilirubin, minimum bilirubin, weight, maximum blood urea nitrogen, and minimum estimated glomerular filtration rate. The logistic regression model performed the best, achieving an AUC score of 0.887 (95% CI: [0.861-0.915]), an accuracy of 0.817, an F1 score of 0.866, a recall score of 0.827, and a Brier score of 0.13.

Conclusion: Compared to the best existing literature in this field, our model achieved an 8.57% improvement in AUC while using 13 less variables, showcasing its effectiveness in determining AKI in septic patients. While the features selected for predicting AKI in septic patients are similar to previous literature, the top features that influenced our model's performance differ.

*Keywords:* Sepsis, Acute Kidney Injury, Machine Learning Models, Prediction, MIMIC-III

## 1. Background

Acute kidney injury (AKI) is an intricate medical condition in which the kidneys are ineffective at filtering waste from the blood. This decline in renal function leads to the accumulation of waste, causing the kidneys to be unable to maintain electrolyte, acid-base, and water balance - all of which can lead to other organ issues such as cardiovascular, gastrointestinal, and neurological complications.[1, 2] Despite the medical advancements in treating AKI, recent studies have found that the mortality rate related to AKI is about 23%.[3]

The most common cause of AKI is sepsis.[4] Sepsis- associated AKI is among the most fatal, with 53% of septic patients developing AKI, and accounts for a mortality rate of up to 44%.[5] Sepsis-associated AKI can also lead to longer stays in the hospital, increase in developing long-term disabilities, and a lower quality of life.[6] Early detection in high-risk septic patients can lead to the reversal of renal failure in early stages, while also improving patient care in the Intensive Care Unit (ICU) and recovery through medical interventions such as Renal Replacement Therapy.[7] It is imperative to understand predictors associated with AKI in septic patients, but the complexity of pathophysiology in sepsis-associated AKI makes it difficult for timely intervention and prevention of renal injury. In general, those who seek medical attention have already developed AKI. Currently, a prominent measure for the diagnosis of AKI is based on increased creatinine levels and a decrease in urine output.[8,4] However, studies have shown that these biomarkers may be nonspecific and require exploring other indicators.[7] When assessing risk factors, a study found that septic shock, hypertension, the use of vasopressors, and mechanical ventilation were among the top determinants that could increase the chance of sepsis-associated AKI.[9]

To mitigate the challenges in the early detection of AKI in septic patients, several studies have investigated various machine learning and statistical methods to construct predictive models.[10] The models utilize specific features or indicators to predict which septic patients develop AKI. In addition to machine learning models, other studies have also used scoring systems such as the simplified acute physiology score (SAPS) II and the sequential organ failure assessment (SOFA), but resulted in poor performance primarily due to low specificity and sensitivity.[5]

Our study contributes to AKI risk prediction by employing a streamlined set of only 23 features, setting our model apart from the extensive variable sets typically used in the models found in literature. Our features were carefully selected to enhance model interpretability and reduce overfitting. This focused feature selection aligns with the best practices in sepsis mortality studies, which prioritize both predictive accuracy and clinical relevance.[11] By adopting a methodology that mirrors rigorous research on sepsis mortality, our approach aims to balance precision with practical usability in clinical settings. Furthermore, the predictive model was developed in accordance with the Transparent Reporting of Individual Prognostic or Diagnostic Multivariate Predictive Model (TRIPOD) guidelines, ensuring transparency and robust cross-cohort comparability.

The structure of our study, from data extraction to validation, mirrors that of recent high-impact research on sepsis mortality, ensuring robust cross-cohort comparability. Additionally, we use Logistic Regression as the primary model due to its balance of performance and interpretability, demonstrating strong results in predictive metrics, including AUC and Brier scores.

In summary, this study advances AKI risk prediction in septic patients by adopting a minimal yet effective feature set of 23 variables and reducing model complexity while maintaining high predictive power. By prioritizing interpretability and alignment with established clinical guidelines (TRIPOD), our approach not only improves transparency but also enhances the model's clinical utility. Logistic Regression was chosen for its balanced performance and ease of interpretation, demonstrating strong predictive metrics like AUC, accuracy, and Brier scores, which support its application in real-world clinical decision-making.

## 2. Methods

*2.1 Data Souce*

The MIMIC-III database, sourced from the ICU at Beth Israel Deaconess Medical Center in Boston, contains extensive health data on over 40,000 patients from 2001 to 2012.[12] This robust dataset features detailed records from more than 58,000 hospitalizations, including physiological signals, medications, lab results, diagnostic data, and medical summaries. It focuses on a diverse range of ICU patients, providing vital insights for clinical decision-making and patient outcomes. The database was crucial for our study as it provided real patient data used to develop our models.

*2.2 Participant Selection*

Our focus was on patients who currently have sepsis. The flowchart in Figure 1 gives a breakdown of the screening and classification requirements done to reach the final dataset. Initially, we filtered the MIMIC-III database using the unique ICD-9 codes of 99591, 99592, and 78552 to uncover patients with sepsis. This left us with 6138 septic patients. We then excluded patients not within the age range of 18 to 89, patients with multiple admissions, patients who stayed in the ICU for less than 48 hours, and patients who had more than 20% of missing values. Lastly, patients with incomplete information were also eliminated. Initially, we encountered 0.7% of our final dataset with more than 11 missing values. Given that this was less than 1% of the data, we removed these rows, assuming it would not affect our result. We finalized the dataset to 3301 septic patients, 2410 of whom developed AKI and 891 who did not.

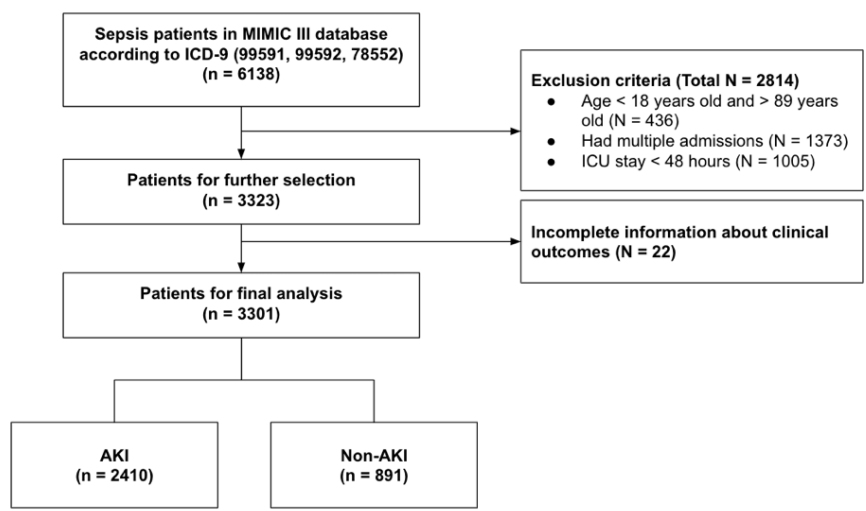

Figure 1: Flow chart of patient selection from MIMIC-III dataset.

*2.3 Data Extraction*

Patient data from the first 24 hours after admission were extracted from the MIMIC-III database. The following information was utilized in this study: (1) demographic characteristics, sex, age, weight, height, and ethnicity; (2) comorbidities, including congestive heart failure, hypertension, cardiac arrhythmias, diabetes, and liver disease; (3) vital signs, including heart rate, temperature, oxygen saturation (SpO2), systolic blood pressure (SysBP), and diastolic blood pressure (DiasBP); (4) laboratory values, including total bilirubin, platelet, anion gap, albumin, chloride, potassium, sodium, lactate, partial thromboplastin time (PTT), prothrombin time (PT), international normalized ratio (INR), creatinine, blood urea nitrogen (BUN), minimum estimated glomerular filtration rate (eGFR), and glucose; and (5) therapeutic and clinical interventions, such as mechanical ventilation and vasopressor use. For variables with multiple measurements, the maximum and minimum values were considered.

To reduce bias caused by missing data, features with more than 20% missing values were excluded from the final cohort. The missing values in other features were imputed using the multiple imputation (MI) method. Multiple Imputation (MI) is a statistical method widely used for handling missing data. It addresses missing values by generating multiple imputed datasets, each with the missing values being replaced with plausible values. These datasets are then analyzed synthetically, obtaining more accurate estimations of the missing values.

*2.4 Feature Selection*

After an initial filtering of septic patients and grouping them into AKI and non-AKI, we started with 50 features. To determine the top features, we used a correlation-based method. We tested to see which features had an absolute value between 0.1 and 1, leaving us with the 23 most correlated features to AKI. All the features that correlate greater than 0.1 or less than -0.1 were considered. Table 1 shows the 23 features selected, which are (i) age: patient age when enter ICU; (ii) minimum heart rate: patient's minimum recorded heart rate; (iii) minimum temperature: patient's minimum recorded temperature; (iv) minimum oxygen saturation (SpO2): patient's minimum recorded oxygen saturation; (v) minimum systolic blood pressure (SysBP): patient's minimum recorded systolic blood pressure; (vi) minimum diastolic blood pressure (DiasBP): patient's minimum recorded diastolic blood pressure; (vii) minimum bilirubin: patient's minimum recorded bilirubin from lab results; (viii) maximum bilirubin: patient's maximum recorded bilirubin from lab results; (ix) minimum anion gap: patient's minimum recorded anion gap from lab results; (x) maximum anion gap: patient's maximum recorded anion gap from lab results; (xi) minimum potassium: patient's minimum recorded potassium level from lab results; (xii) maximum potassium: patient's maximum recorded potassium level from lab results; (xiii) minimum lactate: patient's minimum recorded lactate level from lab results; (xiv) maximum lactate: patient's maximum recorded lactate level from lab results; (xv) minimum creatinine: patient's minimum recorded creatinine level from lab results; (xvi) maximum creatinine: patient's maximum recorded creatinine level from lab results; (xvii) minimum blood urea nitrogen (BUN): patient's minimum recorded blood urea nitrogen from lab results; (xviii) maximum blood urea nitrogen (BUN): patient's maximum recorded blood urea nitrogen from lab results; (xix) urine output: patient's recorded urine output; (xx) vasopressor (vaso): whether patient took vasopressor or not; (xxi) weight: patient's weight at ICU admission; (xxii) minimum estimated glomerular filtration rate (eGFR): patient's minimum recorded estimated glomerular filtration rate from lab results; and (xxiii) mechanical ventilation: whether a patient uses mechanical ventilation or not.

Table 1: Feature Category Table

| Category | Features |
| --- | --- |
| Demographic | Age (years), Weight (lbs) |
| Vital Signs | Minimum Heart Rate (per minute), Minimum Temperature (Celcius), Minimum Oxygen Saturation (SpO2) (%), Minimum Systolic Blood Pressure (SysBP) (mmHg), Minimum Diastolic Blood Pressure (DiasBP) (mmHg), Urine Output (ml) |
| Laboratory Results | Minimum Bilirubin (mg/dL), Maximum Bilirubin (mg/dL), Minimum Anion Gap (mmol/L), Maximum Anion Gap (mmol/L), Minimum Potassium (mEq/L), Maximum Potassium (mEq/L), Minimum Lactate (mmol/L), Maximum Lactate (mmol/L), Minimum Creatinine (mg/dL), Maximum Creatinine (mg/dL), Minimum Blood Urea Nitrogen (BUN) (mg/dl), Maximum Blood Urea Nitrogen (BUN) (mg/dl), Minimum Estimated Glomerular Filtration Rate (eGFR) |
| Interventions | Mechanical Ventilation, Vasopressor (Vaso) |

Our top features were urine output, minimum eGFR, maximum eGFR, minimum SysBP, mechanical ventilation, vasopressor, weight, and minimum creatinine. Existing literature does include urine output, mechanical ventilation, vasopressor, and creatinine as top features, but the high correlation of minimum eGFR, maximum eGFR, minimum SysBP, and weight for predicting AKI in septic patients is unique to our model.

Urine output is one of the most common signs of AKI. Studies have been done to show the ideal amount and time span to collect urine sample measurements.[13] A decrease in urine content is typically a strong sign of AKI, making its feature importance high in prediction. Regarding eGFR levels, it has been seen to decrease after patients are diagnosed with AKI.[14] While there is still ongoing research regarding the effect of eGFR and AKI, it is a notable feature that should be looked at for prediction. Systolic blood pressure is another indicator that should be looked at when determining an AKI diagnosis because based on previous studies, it is vital for organ perfusion, where blood flow provides oxygen and nutrients to the organs.[15,16] Vasopressor is a drug used to increase blood pressure. It can be used to regulate hypotension by constricting blood vessels.[17] In our dataset, it was recorded whether a patient took vasopressor or not. Weight is another significant factor in predicting AKI. Studies have stated that obesity is associated with an increased risk of AKI in areas such as acute respiratory distress syndrome or post cardiac surgery.[18,19] In a recent study involving critically ill septic patients in the ICU, it was found that patients with obesity had a higher chance of developing early sepsis-associated AKI.[20] The literature supports our findings of the features that will be used in developing a model to predict AKI in septic patients.

*2.5. Statistical Analysis for AKI and Non-AKI Patients*

To compare the features for AKI and non-AKI patients, we utilized statistical tests with an alpha set to 0.05 to compare the two cohorts and see if there is any statistically significant differences. The results can be found in Table 2. The quantitative values were compared using a T-Test, and the qualitative values were compared using a chi-squared test. All features were statistically significant between AKI and non-AKI patients.

*2.6. Model Development*

Our filtered dataset was imbalanced between AKI patients and non-AKI patients. There were 2410 AKI patients, and 891 non-AKI patients. To fix this, we utilized the Synthetic Minority Over-Sampling Technique (SMOTE), which creates additional data from the minority class using a K Nearest Neighbors approach.[21] SMOTE was added to every feature that has been selected to balance the data and prevent overfitting. The data was then split into a training, testing, and validation cohort. The six models used in this study were Logistic Regression (LR), XGBoost, K Nearest Neighbors (KNN), Support Vector Machines (SVM), Random Forest (RF), and LightGBM.

Logistic regression has been studied to have comparable performance results when compared to machine learning models and has been used in studies for predicting AKI.[22,23] For its parameters, we used a C of 100, a maximum iteration of 200, and a penalty of 'l2'. XGBoost has been used in previous studies for predicting AKI in patients with specific prior health conditions.[24,25] For the parameters, we used the 'binary:logistic' objective, a 'reg lambda' of 100, a 'reg alpha' of 120, and a 'max depth' of 2. KNN is another machine learning algorithm that utilizes classification and has been used for medical predictions, such as predicting development of AKI for heart failure patients.[26,27] For its parameters, we used an n neighbors of 40. SVM is another well-known machine learning model known to provide high accuracy.[28] For SVM, we used a C of 0.1 and a gamma of 0.02. Random forest is a widely utilized classification machine learning model that has been used in several acute kidney injury prediction studies.[29,30] We used an n estimator of 150, a max depth of 12, a min samples split of 128, and a min samples leaf of 10. LightGBM is another commonly used machine learning model used in literature for predicting AKI in septic patients. We used a lambda_l1 of 1,a lambda_l2 of 1, a num_leaves of 20, a max depth of 4, a random state of 42, a learning rate of 0.01, a n estimators of 1050,and a verbose of -1.

Table 2: Characteristics of AKI and Non-AKI cohorts with statistical analysis results.

| Category | AKI (N = 2410) | Non-AKI (N = 891) | P-value |
| --- | --- | --- | --- |
| Demographic | | | |
| Age (years) | 65.849 | 61.700 | <0.0001 |
| Weight (lbs) | 86.547 | 75.287 | <0.0001 |
| Vital Signs | | | |
| Minimum Heart Rate (per minute) | 61.464 | 65.157 | <0.0001 |
| Minimum Temperature (Celsius) | 35.390 | 35.720 | <0.0001 |
| Minimum Oxygen Saturation (SpO2) (%) | 78.595 | 83.512 | <0.0001 |
| Minimum Systolic Blood Pressure (SysBP) (mmHg) | 68.973 | 77.726 | <0.0001 |
| Minimum Diastolic Blood Pressure (DiasBP) (mmHg) | 31.680 | 35.630 | ¡0.0001 |
| Urine Output (ml) | 1189.225 | 2690.867 | <0.0001 |
| Laboratory Results | | | |
| Minimum Bilirubin (mg/dL) | 2.856 | 1.749 | <0.0001 |
| Maximum Bilirubin (mg/dL) | 3.356 | 2.023 | <0.0001 |
| Minimum Anion Gap (mmol/L) | 14.025 | 12.583 | <0.0001 |
| Maximum Anion Gap (mmol/L) | 18.167 | 16.499 | <0.0001 |
| Minimum Potassium (mEq/L) | 3.782 | 3.612 | <0.0001 |
| Maximum Potassium (mEq/L) | 4.758 | 4.440 | <0.0001 |
| Minimum Lactate (mmol/L) | 1.942 | 1.581 | <0.0001 |
| Maximum Lactate (mmol/L) | 3.551 | 2.821 | <0.0001 |
| Minimum Creatinine (mg/dL) | 1.914 | 1.252 | <0.0001 |
| Maximum Creatinine (mg/dL) | 2.366 | 1.639 | <0.0001 |
| Minimum Blood Urea Nitrogen (BUN) (mg/dL) | 35.783 | 27.436 | <0.0001 |
| Maximum Blood Urea Nitrogen (BUN) (mg/dL) | 42.460 | 34.530 | <0.0001 |
| Minimum Estimated Glomerular Filtration Rate (eGFR) | 62.955 | 82.275 | <0.0001 |
| Interventions | | | |
| Vasopressor (Vaso) | 1840 (count) | 521 (count) | <0.0001 |
| Mechanical Ventilation | 1500 (count) | 347 (count) | <0.0001 |

The best model was determined based on its respective AUC value. While we did also look at the accuracy, F1 score, and recall of each model, we decided to use AUC as the metric of comparison since it is more suitable at handling imbalanced datasets and when comparing multiple models.

For this study, the data was imported to BigQuery and extracted using SQL. The data preprocessing, feature selection, and training was done using Python 3.10.12.

*2.7. Shapley Analysis*

To interpret the predictive power of each feature within our model, we employed Shapley Analysis. Shapley values assign a contribution score to each feature based on its marginal impact on the model's output, taking into account all possible feature combinations. This approach provides a comprehensive measure of feature importance that is both rigorous and interpretable, making it especially suited to healthcare applications where understanding each variable's influence is crucial. Shapley Analysis helps us identify the most critical predictors, such as urine output, bilirubin, and blood urea nitrogen levels. These values are particularly useful for determining how each feature individually, as well as in combination with others, influences the likelihood of AKI, thus allowing for more informed clinical insights.

Analysis was chosen due to its capacity to illustrate feature contributions transparently, aligning with the need for interpretability in clinical decision-making tools.[31] This analysis enabled us to pinpoint the most impactful features, visualized in our results section, that support our model's application in real-world settings where clinicians can focus on key indicators of AKI risk.

## 3. Results

*3.1. Training and Validation Comparison*

After an initial filtering of patients who did not meet requirements for analysis, we were left with 3301 patients and 56 features. After narrowing down our features to 23 using correlation, the data was then randomly divided into 40% for the training set, 10% for the test set, and 50% for the validation set, resulting in 1980 patients for training, 661 patients for testing, and 660 for validation. Table 3 gives a comprehensive comparison between the training and validation sets. The comparison of the training and validation set is to confirm that our null hypothesis of no difference between feature distributions holds.

*3.2. Model Performance Evaluation*

After running each of the six models, we have documented our results in Tables 4 and 5. The Logistic Regression model had a training set AUC of 0.898 (95% CI = [0.888 - 0.909]), a validation set AUC of 0.857 (95% CI = [0.829 – 0.885]), and a testing set AUC of 0.887 (95% CI = [0.861 - 0.915]). Regarding the other five models, LightGBM had an AUC of 0.885, SVM 0.876, random forest 0.867, XGBoost 0.862, and KNN 0.725.

Figure 2 shows the AUC curves of each of the 6 models tested. Logistic Regression exhibits the closest shape to an ideal AUC curve, showcasing its superior performance amongst the other models.

Table 3: Averages and P-values between features of the training and validation set.

| Category | Training Set (N = 1980) | Validation Set (N = 660) | P-value |
|---|---|---|---|
| **Demographic** | | | |
| Age (years) | 64.8 | 64.1 | 0.303 |
| Weight (lbs) | 83.4 | 84.0 | 0.647 |
| **Vital Signs** | | | |
| Minimum Heart Rate (per minute) | 62.7 | 62.1 | 0.358 |
| Minimum Temperature (Celsius) | 35.5 | 35.5 | 0.548 |
| Minimum Oxygen Saturation (SpO2) (%) | 80.0 | 80.0 | 0.946 |
| Minimum Systolic Blood Pressure (SysBP) (mmHg) | 71.1 | 72.2 | 0.198 |
| Minimum Diastolic Blood Pressure (DiasBP) (mmHg) | 32.6 | 33.1 | 0.290 |
| Urine Output (ml) | 1576 | 1696 | 0.085 |
| **Laboratory Results** | | | |
| Minimum Bilirubin (mg/dL) | 2.62 | 2.51 | 0.627 |
| Maximum Bilirubin (mg/dL) | 3.05 | 3.00 | 0.823 |
| Minimum Anion Gap (mmol/L) | 13.6 | 13.6 | 0.936 |
| Maximum Anion Gap (mmol/L) | 17.7 | 17.7 | 0.796 |
| Minimum Potassium (mEq/L) | 3.73 | 3.73 | 0.920 |
| Maximum Potassium (mEq/L) | 4.66 | 4.66 | 0.907 |
| Minimum Lactate (mmol/L) | 1.84 | 1.86 | 0.717 |
| Maximum Lactate (mmol/L) | 3.36 | 3.32 | 0.767 |
| Minimum Creatinine (mg/dL) | 1.73 | 1.72 | 0.885 |
| Maximum Creatinine (mg/dL) | 2.17 | 2.15 | 0.764 |
| Minimum Blood Urea Nitrogen (BUN) (mg/dL) | 33.7 | 32.9 | 0.453 |
| Maximum Blood Urea Nitrogen (BUN) (mg/dL) | 40.6 | 39.7 | 0.488 |
| Minimum Estimated Glomerular Filtration Rate (eGFR) | 68.0 | 69.0 | 0.520 |
| **Interventions** | | | |
| Vasopressor (Vaso) | 1425 (count) | 465 (count) | 0.485 |
| Mechanical Ventilation | 1104 (count) | 367 (count) | 0.981 |

Table 4: Logistic Regression Model training, testing, and validation sets evaluation metrics.

| Models | AUC | AUC 95% CI | Accuracy |
|---|---|---|---|
| Proposed model performance on test set | 0.887 | [0.861-0.915] | 0.817 |
| Proposed model performance on validation set | 0.857 | [0.829–0.885] | 0.791 |
| Proposed model performance on training set | 0.898 | [0.888-0.909] | 0.819 |

Table 5: Model Evaluation Metrics

| Models | AUC | AUC 95% CI | Accuracy | F1 Score | Recall |
| --- | --- | --- | --- | --- | --- |
| Logistic Regression | 0.887 | [0.861-0.915] | 0.817 | 0.867 | 0.827 |
| LightGBM | 0.885 | [0.860-0.911] | 0.815 | 0.869 | 0.853 |
| SVM | 0.876 | [0.844-0.904] | 0.797 | 0.849 | 0.794 |
| Random Forest | 0.867 | [0.839-0.895] | 0.790 | 0.848 | 0.819 |
| XGBoost | 0.862 | [0.834-0.891] | 0.796 | 0.777 | 0.739 |
| KNN | 0.725 | [0.683-0.764] | 0.696 | 0.777 | 0.739 |

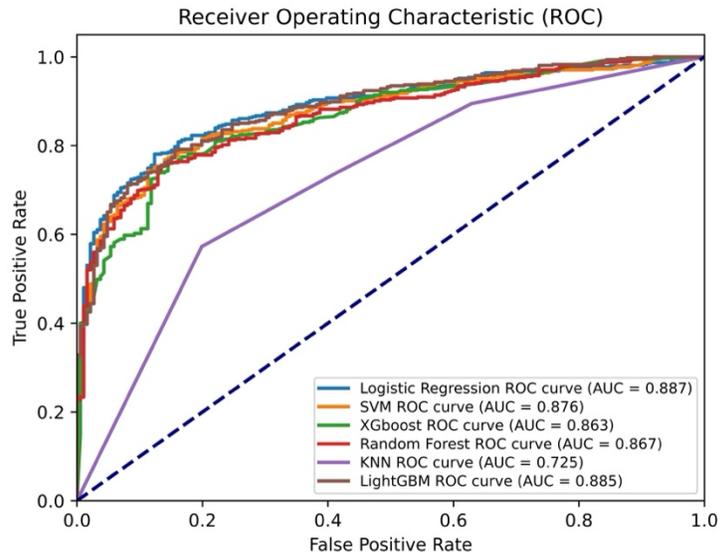

Figure 2: ROC curves of Baseline and Proposed models.

In addition, we conducted calibration techniques and rigorous tests for each model. The Brier Score measured the mean squared difference between predicted probabilities and actual outcomes, with lower scores indicating better calibration. Isotonic Regression, a non-parametric method, was used for calibration. Figure 3 indicates that the predicted probabilities are well-calibrated, as the points are positioned near the diagonal line. The results of Brier scores are as follows: Logistic Regression 0.134; LightGBM 0.144; SVM 0.141; Random Forest 0.143; XGBoost 0.147; and KNN 0.229. Of all the models, the best-performing model according to the Brier score was Logistic regression (0.134), indicating a good calibration. The low Brier Score and high AUC support its accuracy and reliability.

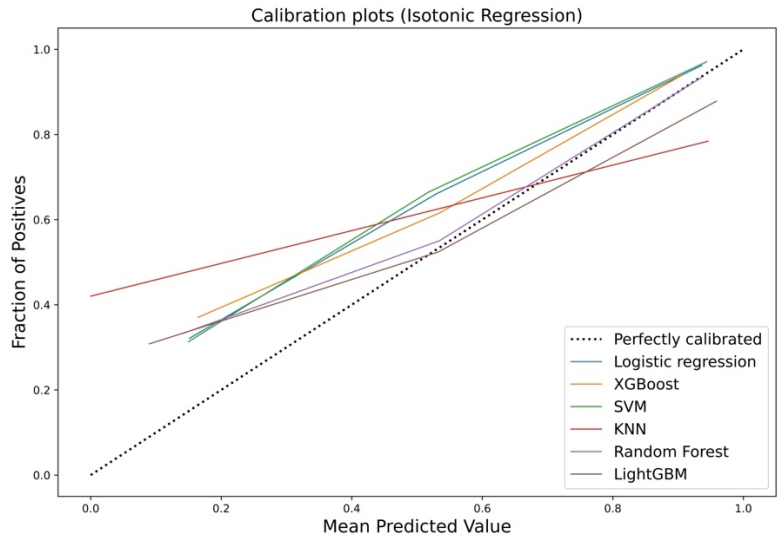

Figure 3: Calibration curves of the 6 models used: Logistic Regression, SVM, XGBoost, Random Forest, KNN, and LightGBM.

*3.3. Shapley Analysis for Logistic Regression*

Shapley analysis is a common measure to see which features contribute the most to the model's overall performance. Since Logistic Regression was our best model, we applied Shapley analysis to see which features most affected the outcome of our testing set.

Figure 4 visualizes the Shapley analysis for our Logistic Regression model. Urine output, maximum bilirubin, minimum bilirubin, and weight are the most important features used in the Logistic Regression prediction model. From Figure 5, we see that urine output has the most significant impact on the Logistic Regression model. Low urine output increases the risk of AKI, displayed by its negative SHAP value. This is consistent with clinical understanding where reduced urine output is a key indicator of acute kidney injury (AKI).[32] Elevated bilirubin levels push predictions toward higher risk, reflecting liver dysfunction or sepsis-induced cholestasis, conditions often seen in critical care patients.[33]

When comparing to the original feature importance, it is interesting to note that bilirubin now plays an important role in the predictive process. While in the original feature correlation results it was not among the top features, it now holds a higher predictive power in the Logistic Regression model. This highlights the importance of performing additional analysis to understand what specific features the model looks at to make predictions.

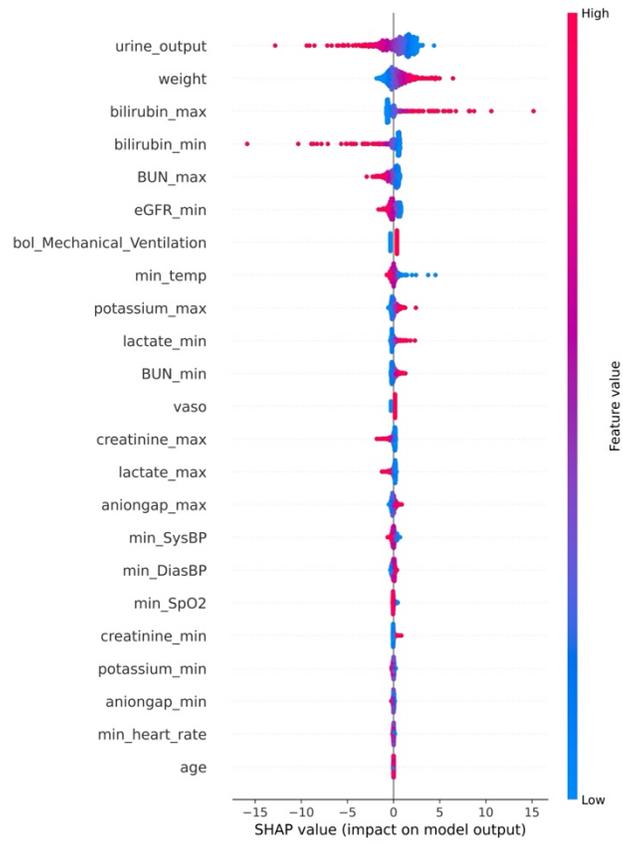

Figure 4: Shapley Analysis of Logistic Model features.

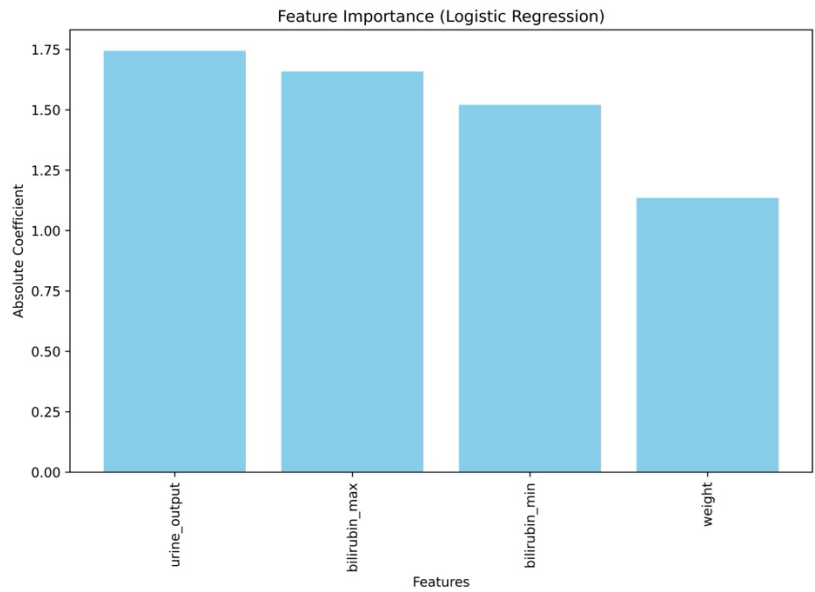

Figure 5: Feature importance for logistic regression test set. The predictors are urine output, maximum bilirubin, minimum bilirubin, and weight.

# 4. Discussion

*4.1. Existing Literature Comparison*

Our model aims to predict acute kidney injury in septic patients. Among all the models we have trained, the proposed Logistic Regression model shows an impressive outcome with an AUC of 0.887, which is only 0.23% higher than the second best model trained, LightGBM. These results exceed the general pattern seen in previous literature.

In the field of predicting acute kidney injury, several studies have generated proper outcomes using different methods, including XGBoost, LightGBM, and Nomogram. [5,34-37] Compared to the best literature model[5], our model shows a significant improvement. The best literature model uses XGBoost, while ours uses Logistic Regression. Our model shows an 8.56% improvement in AUC. Additionally, our AUC values contain precise confidence intervals, ensuring reliability of our results. While the best existing model used 36 variables, our model was able to use 13 less variables and achieve a higher AUC score. The lower feature count prevents issues such as overfitting and highlights the efficiency of our model. This is crucial when working with sparse medical data to ensure accurate predictions. Like the best existing literature, we also utilized a training, validation, and test set. Regarding the features used, we have three features not included in the model of the current best literature: weight, minimum heart rate, and minimum oxygen saturation (SPO2). After using Shapley analysis on our Logistic Regression model, it highlighted new top features that were not discussed in previous literature, such as bilirubin and maximum blood urea nitrogen. This information could be helpful to healthcare workers to see which patient features they need to monitor and evaluate.

While logistic regression exhibited the best performance, LightGBM, SVM, Random Forest, and XGBoost showed similar performance levels based on their AUC value and confidence intervals, as seen in Figure 6. This could mean that other models, when using accurate features and hyperparameters, are also able to accurately predict AKI in septic patients.

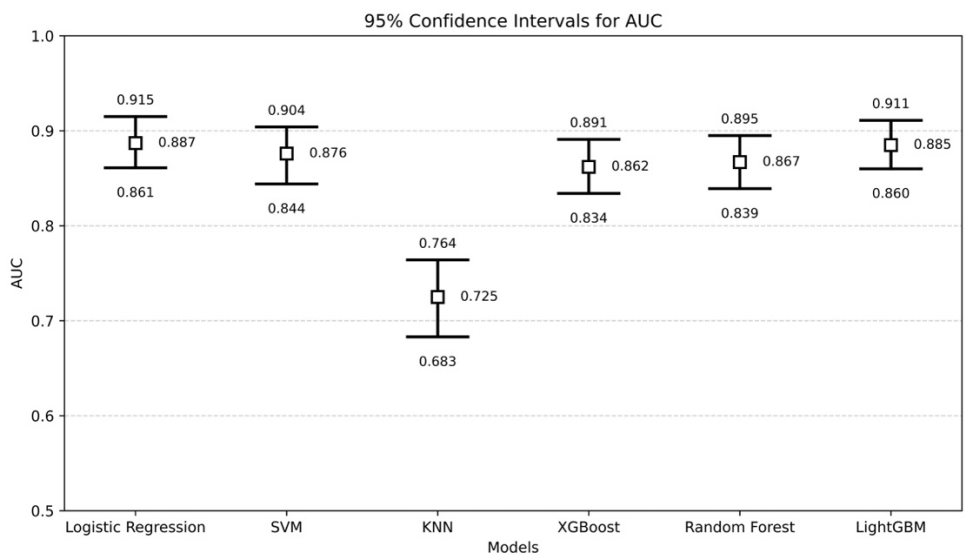

Figure 6: AUC with Confidence Intervals for each Model.

*4.2. Study Limitations*

While the model shows a satisfactory result, some limitations cannot be overlooked. The data in the MIMIC-III database comes from the Beth Israel Deaconess Medical Center, a tertiary academic medical center located in Boston, Massachusetts. Using medical data from one single medical center might lead to biased predictions in some particular areas like ethnicity, which were excluded during the process of our feature selection. Therefore, incorporating multiple databases would likely enhance the reliability of predictions by providing a more diverse and representative sample. [38] Meanwhile, we cannot lose sight of the importance of ethnicity in the field of biological and medical research. Hence, some further improvements can be made by eliminating the bias caused by using a singular dataset. One study concluded that the diverse populations, results, and features are hindering the implementation of predictive models to aid septic patients with AKI.[39] The heterogeneity of medical patient data available will need to be addressed to improve the model's performance and reliability going forward.

*4.3. Future Works*

To enhance the clinical utility and robustness of AKI prediction, future research should test the model across diverse healthcare settings, expanding beyond the MIMIC-III database to improve generalizability. Future work could involve testing our model on diverse medical datasets to see if the results hold, providing a more inclusive model that considers potential demographic disparities. Additionally, integrating real-time monitoring could allow for continuous risk assessment, adapting to dynamic changes in patient conditions during ICU stays. Lastly, exploring advanced modeling techniques, such as ensemble learning or deep learning architectures, may offer further predictive power while maintaining interpretability for practical application. As more studies between sepsis and AKI are conducted, researchers could also test out new features to add or remove from the model to see if its performance increases. Furthermore, fine tuning the hyperparameters of our current model could also lead to improved AUC scores.

## 5. Conclusion

Our study was able to produce a well-trained model aimed at predicting acute kidney injury in septic patients using multiple machine learning algorithms and advanced data preprocessing techniques, which include reducing bias by excluding features with a high percentage of missing values and implementing multiple imputation to handle missing data from the MIMIC-III database. Our model is distinguished by its high AUC compared to the other five baseline models, and its stability as seen through the narrow confidence interval. Compared to previous literature, our model is able to achieve higher AUC while using a limited number of features. For implementation in the medical field, a model with fewer features allows for a more concise and comprehensive review of patient data in a shorter amount of time. For determining feature importance, we utilized Shapley analysis, a method known for evaluating the predictive power of each feature in machine learning. It allowed us to identify the most important predictors, which were categorized into the following groups: demographic, vital signs, laboratory results, and interventions. Based on our findings, urine output, bilirubin, and weight were the most valuable predictors in determining AKI in septic patients. This information can be used to assist medical professionals at promptly identifying patients more at risk for developing AKI.

Our research has provided a new model aimed at predicting AKI in septic patients, using existing data from MIMIC-III, which includes demographic, laboratory, vital signs, and intervention data. From there, we used feature engineering to select the most relevant features for AKI prediction in septic patients, and cross-validated our results to existing literature, ensuring our model is accurate.

Machine learning models can be a vital resource for the medical field in early detection and intervention for septic patients who exhibit AKI symptoms. The model can assist medical professionals with early diagnosis of AKI, reducing the mortality rate related to sepsis-associated AKI. This also aids with the decision making process for medical professionals as they are able to assess which patients need immediate medical attention. While our model showed superior performance, SVM, XGBoost, Random Forest, and LightGBM were promising models, indicating the accuracy and reliability of machine learning in the medical field. While clinicians should not exclusively rely on machine learning models to make decisions, it can be used as a reference in conjunction with their expert knowledge.

Our study showcased the value of utilizing machine learning in the medical field. Our logistic regression model displayed superior performance in predicting AKI in septic patients, with an AUC of 0.887 (95% CI: [0.861-0.915]) and an accuracy of 0.817, while utilizing only 23 features, 13 less than current literature. Our model could assist medical professionals in evaluating patients, making diagnoses, and determining the appropriate next steps for treatment.

**Acknowledgments**

The authors extend their gratitude to the creators of MIMIC-III for furnishing a thorough and inclusive public electronic health record (EHR) dataset.[40]

**Author Contributions**

Aleyeh Roknaldin: Conceptualization, Methodology, Validation, Visualization, Writing – original draft, Supervision. Zehao Zhang: Formal analysis, Software, Writing – review editing, Visualization, Data curation. Jiayuan Xu: Writing – review editing, Conceptualization, Methodology, Data curation. Maryam Pishgar: Writing – review editing, Validation, Supervision, Software, Resources, Project administration, Investigation, Funding acquisition, Formal analysis, Data curation, Conceptualization.

**Statements and Declarations**

Not Applicable

**Ethical Considerations**

The predictive model was achieved based on guidelines of the Transparent Reporting of Individual Prognostic or Diagnostic Multivariate Predictive Model (TRIPOD) initiative.

**Consent to Participate**

Not Applicable

**Consent for Publication**

Not Applicable

**Declaration of Conflicting Interest**

The authors declare that they have no known competing financial interests or personal relationships that could have appeared to influence the work reported in this paper.

**Funding Statement**

No funding was used for the ideation, execution, or writing of this study.

**Data Availability**

The data was sourced from the MIMIC-III database, which is not publicly available due to privacy, but access is available through the corresponding author upon reasonable request.